\documentclass{article}
\usepackage{spconf,amsmath,graphicx}
\usepackage{epsfig}
\usepackage{graphicx}
\usepackage{amsmath}
\usepackage{amssymb}
\usepackage[retainorgcmds]{IEEEtrantools}
\usepackage{algorithmic}
\usepackage{algorithm}
\usepackage{hyperref}
\usepackage[applemac]{inputenc}
\usepackage{pst-sigsys}

\graphicspath{{/Users/joan/Documents/papers/figures/}}

\providecommand{\argmin}{\mathop{\textup{argmin}}}

\newcommand{\ga}{\gamma}

\newcommand{\cC}{{\cal C}}

\newcommand{\wavel}{W_{j,\gamma}}
\newcommand{\oavel}{\overline{W}_J}
\newcommand{\pavel}{\overline{U}_J}

\title{Classification with Invariant Scattering Representations}
\name{Joan Bruna and Stéphane Mallat}
\address{CMAP, École Polytechnique\\ Rue de Saclay, Palaiseau, France}
\begin{document}
%
\maketitle

\begin{abstract}
A scattering transform defines a signal representation
which is invariant to translations and Lipschitz continuous
relatively to deformations. It is implemented with a non-linear
convolution network that iterates over wavelet and modulus operators.
Lipschitz continuity locally linearizes deformations.
Complex classes of signals and textures can be modeled with 
low-dimensional affine spaces, computed
with a PCA in the scattering domain. Classification is performed
with a penalized model selection.
State of the art results are obtained for handwritten digit
recognition over small training sets, and for texture classification.
\footnote{This work is funded by the ANR grant 0126 01.}
\end{abstract}


\section{Introduction}
Affine space models are simple to compute with a Principal Component Analysis 
(PCA) but are not appropriate to
approximate signal classes that include complex forms 
of variability. Image classes are often invariant to rigid transformations 
such as translations or rotations, and
include elastic deformations, which define highly non-linear manifolds.
Textures may also be
realizations of strongly non-Gaussian processes that cannot
be discriminated with linear models either.

Kernel methods define 
distances $d(f,g) = \|\Phi(f) - \Phi(g) \|$, with operators
$\Phi$ which address these issues by 
mapping $f$ and $g$ into a space of much higher
dimension. However,
invariance properties and learning
requirements on small training sets, rather suggest to implement a
dimensionality reduction. 


Scattering operators constructed in \cite{stephane0,stephane1}, 
are invariant to global translations and 
Lipschitz continuous relatively to local deformations, up to a log term, 
thus providing local translation invariance through the linearization
of such deformations.
These scattering operators create invariance by averaging
interference coefficients, which capture signal interactions at several
scales and orientations.
This paper models complex signal classes with low-dimensional affine spaces
in the scattering domain, which are computed with a PCA.
The classification is performed by a penalized model selection.

Scattering operators may also be invariant to any
compact Lie subgroup of $GL(\mathbb{R}^2)$, such as rotations,
but we concentrate on translation invariance,
which carries the main difficulties and already covers a wide range of
classification applications. 
Section \ref{sec2} reviews the construction of scattering operators 
with a cascade of wavelet transforms and modulus operators, which defines a
non-linear convolution network \cite{LeCun}.
Section \ref{section_classif} shows that
learning affine scattering model spaces has 
a linear complexity in the number of training samples.
Section \ref{sec_num_results} describes state of the art classification
results obtained from limited number of training samples
in the MNIST hand-written digit database, and 
for texture classification in the CUREt database.
Software is available at \texttt{ www.cmap.polytechnique.fr/scattering}.

\section{Scattering Operators}\label{sec2}

In order to build a representation which is locally 
translation invariant,
a scattering transform begins from a wavelet representation. Translation
invariance is obtained by progressively mapping high frequency wavelet
coefficients to lower frequencies, with modulus operators described 
in Section \ref{Waveletmodulu}.
Scattering operators, defined in \ref{scaopern}, 
iterate over wavelet modulus operators. 
Section \ref{scatmetric} shows that it defines a translation
invariant representation, which is Lipschitz continuous to deformation, up
to a log term.

\subsection{Wavelet Modulus Propagator}
\label{Waveletmodulu}

A wavelet transform
extracts information at different scales and orientations
by convolving a signal $f$ 
with dilated bandpass wavelets $\psi_\gamma$ having a 
spatial orientation angle $\gamma \in \Gamma$:
\[
W_{j,\gamma}f(x) = f \star \psi_{j,\gamma}(x)~\mbox{with}~\psi_{j, \gamma}(x) = 2^{-2j}\psi_\gamma(2^j x).
\] 
At the largest scale $2^J$, low-frequencies are carried by 
a lowpass scaling function $\phi$: 
$A_J f = f \star \phi_J$, with $\phi_J(x) = 2^{-2J}\phi(2^{-J} x)$
and $\int \phi(x)\,dx = 1$. 
The resulting wavelet representation is 
$$\overline{W}_Jf = \{A_J f,\, \wavel f\}_{j < J,\gamma \in \Gamma}.$$
The norm of the wavelet operator is defined by
\begin{equation}\label{normwaveoperator}
\|\overline{W}_J f\|^2 = \|f \star \phi_J\|^2 + \sum_{j<J,\gamma \in \Gamma} \| \wavel f\|^2
\end{equation}
with $\|f\|^2 = \int |f(x)|^2\,dx$ and it satisfies
\begin{equation}\label{littlewood1}
(1-\delta) \|f\|^2 \leq \|\overline{W}_J f\|^2 \leq \|f\|^2
\end{equation}
if and only if for all $\omega \in \mathbb{R}^2$,
\small
\begin{equation}\label{littlewood2}
1 - \delta \leq |\hat{\phi}(2^J\omega)|^2 + \frac{1}{2} \sum_{j<J,\gamma \in \Gamma} \left( |\hat{\psi}_\gamma(2^j \omega)|^2 + |\hat{\psi}_\gamma(-2^j \omega)|^2\right) \leq 1.
\end{equation}
\normalsize

We consider families of complex wavelets 
$$\psi_\gamma(x) = e^{i \xi_\gamma x} \theta_\gamma(x) $$ where
$\theta_\gamma (x)$ are low-pass envelops.
Oriented Gabor functions are examples of complex wavelets, obtained 
with a modulated Gaussian $\psi(x) = e^{i\xi \cdot x} e^{-|x|^2/(2\sigma^2)}$,
which is rotated with $R_\gamma$ by an angle $\pi \gamma / |\Gamma|$:
$\psi_\gamma(x) = \psi(R_\gamma x)$. In numerical experiments, we
set $\xi = 3 \pi / 4$, $\sigma = 1$, $|\Gamma| = 6$, and
$\phi$ is also a Gaussian with $\sigma = 2/3$. It satisfies
(\ref{littlewood2}) only over a finite range of scales.

If $f_\tau(x)=f(x-\tau)$ then
\[
W_{j,\gamma} f_\tau(x) = W_{j,\gamma} f(x-\tau) \approx W_{j,\gamma}f(x)
\]
if and only if $|\tau| \ll 2^j$, because
$ W_{j,\gamma} f$ has derivatives of amplitude
proportional to $2^{-j}$. High frequencies
corresponding to fine scales are thus highly sensitive to translations.

Translation invariance is improved by mapping
high frequencies to lower frequencies with a complex modulus operator.
Since $\psi_\gamma(x) = e^{i \xi_\gamma x} \theta_\gamma(x)$,
we verify that
$$\wavel f(x) = e^{i \xi_{j,\gamma}x} \left( f_{j,\gamma} \star \theta_{j,\gamma}(x)\right),$$
where $\xi_{j,\gamma}=2^{-j}\xi_\gamma$,
$ \theta_{j,\gamma}(x) = 2^{-2j} \theta_\gamma(2^{-j}x)$ and
$f_{j,\gamma}(x) = e^{i \xi_{j,\gamma}x}f(x)$. 
Wavelet coefficients $\wavel f(x)$ are located at high frequencies because
of the $e^{i \xi_{j,\gamma}x} $ term. 
These oscillations are removed by a modulus operator
\begin{equation}\label{modulus}
|\wavel f| = |  f_{j,\gamma} \star \theta_{j,\gamma}(x) |~.
\end{equation}
The energy of $|\wavel f|$
 is now mostly concentrated in the low frequency domain
 covered by the envelop $\hat{\theta}_{j,\gamma}(\omega) = \hat{\theta_\gamma}(2^j \omega)$. 
 It may however also include some high frequencies produced by
the modulus singularities where $  f_{j,\gamma} \star \theta_{j,\gamma}(x) =0$.
Using complex wavelets is important to reduce the number of such singularities
and thus concentrate the information at low frequencies.

If $f(x) = \sum_n a_n \cos (\omega_n x)$ then one can verify
that 
$|W_{j,\gamma} f(x)| = c_{j,\gamma} + \epsilon_{j,\gamma} (x)$ where
$\epsilon_{j,\gamma}(x)$ is an interference term.
It is a combination of the 
$\cos (\omega_n - \omega_{n'})x $, for all $\omega_n$ and $\omega_{n'}$ 
in the support of $\hat \psi_\gamma (2^j \omega)$.
The modulus yields interferences
that depend upon frequency intervals, but
it loses the exact frequency locations
$\omega_n$ in each octave.

A wavelet modulus propagator
is obtained by applying a 
complex modulus to all wavelet 
coefficients:
$$\pavel f = \{ A_J f, |\wavel f|\}_{j<J, \gamma \in \Gamma}~.$$
Since $||a| - |b|| \leq |a-b|$ and the wavelet transform is contractive,
it results that
\[
\|\pavel f - \pavel g \| \leq \|\oavel f - \oavel g\| \leq \| f-g\| 
\]
and $\|\pavel f\| = \|f\|$ if $\delta = 0$ in (\ref{littlewood1}).


\subsection{Multiple Paths Scattering}
\label{scaopern}

Thanks to the concentration towards
the low frequencies, the interference
coefficients of the wavelet modulus propagator
can be locally averaged by $\phi_J$ in order
to produce locally translation invariant coefficients 
with non negligible energy:

\[
|f \star \psi_{j,\gamma}| \star \phi_J (x).
\]

The high frequencies of $|f \star \psi_{j_1,\ga_1}|$ not removed by the 
convolution with
$\phi_J$ are carried by the wavelet coefficients $|f \star \psi_{j_1,\ga_1}| 
\star \psi_{j_2,\ga_2}$ at scales $2^{j_2} < 2^{J}$.
To become insensitive to local translation and reduce the variability
of these coefficients, their complex phase can also be removed by a modulus which is
also averaged by $\phi_J$:
\[
||f \star \psi_{j_1,\ga_1}| \star \psi_{j_2,\ga_2}| \star \phi_J~.
\]
These second order coefficients provide co-occurrence information at 
two scales $2^{j_1}$, $2^{j_2}$ and in two directions ${\ga_1}$ and
$\ga_2$. This can distinguish corners and junctions from edges and characterize
texture structures.
Coefficients are only calculated for $2^{j_2} < 2^{j_1}$ because one can show
\cite{stephane1} that if $\psi$ is an appropriate complex wavelet then
$|f \star \psi_{j_1,\ga_1}|\star \psi_{j_2,\ga_2}$
is negligible at scales $2^{j_2} \geq 2^{j_1}$. 

The high frequencies lost by the filtering with $\phi_J$ can again be restored with
finer scale wavelet coefficients, which are regularized by suppressing their phase
with a modulus and by
averaging the result with $\phi_J$. Applying iteratively this procedure
$n$ times yields the following coefficients:
\[
|||f \star \psi_{j_1,\ga_1}| \star \psi_{j_2,\ga_2}| ... \star \psi_{j_n,\ga_n}|\star \phi_J(x)~.
\]
At any location $x$, they provide co-occurrence information between any of the
$|\Gamma|^n$ families of angles $1 \leq \ga_1,...,\ga_n \leq |\Gamma|$ 
and any of the $\binom{J}{n}$ families
of scales satisfying $0 \leq j_1 < ...<j_n < J$. 
They are called scattering
coefficients because they can be interpreted as interaction coefficients
between  $f$ and the successive wavelets $\psi_{j_1,\ga_1}\,$...$\,\psi_{j_n,\ga_n}$.

A scattering operator considers all the 
scattering coefficients at all scales and orientations up
to a maximum co-occurrence order $m$.
It is indexed  along a path variable 
$p=\{(j_n,\gamma_n) \}_{ n\leq |p| \leq m}$ which is a family of wavelet
indices. 
It computes $|p|$ wavelet convolutions and modulus along the 
path
$$S_J (p)f = \underbrace{| \cdots |}_{|p|}  f \star \psi_{j_1,\gamma_1} | \star \psi_{j_2,\gamma_2}| \dots | \star \psi_{j_{|p|},\gamma_{|p|}} | \star \phi_J  $$ 
with $  j_n < J$ and $\gamma_n \in \Gamma$. Its dimension is $\sum_{n=0}^m |\Gamma|^n\binom{J}{n}$.


One can verify that scattering coefficients for paths of length 
$m'$ are computed
by applying the wavelet modulus propagator $\pavel$
to scattering coefficients for all paths $p$ of length $|p| = m'-1$: 
\begin{equation}\label{cascadeop}
\{\pavel S(p)f\}_{p,|p|=m'-1} =\{ S_J(p) f\}_{p,|p|=m'-1} \cup \{S(p) f\}_{p,|p|=m'}~.
\end{equation}
where $S (p)f = \underbrace{| \cdots |}_{|p|}  f \star \psi_{j_1,\gamma_1} | \star \psi_{j_2,\gamma_2}| \dots | \star \psi_{j_{|p|},\gamma_{|p|}} |$.

A scattering operator is thus computed with a cascade of convolutions
and modulus operators over $m+1$ layers,
similar to the convolution network architecture introduced 
by LeCun \cite{LeCun,Poggio}: 
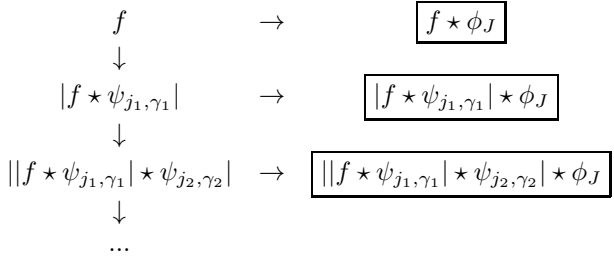
\begin{equation*}
\begin{matrix}
f & \rightarrow & \boxed{f \star \phi_J} \\
\downarrow & & \\
|f \star \psi_{j_1,\ga_1}| & \rightarrow & \boxed{|f \star \psi_{j_1,\ga_1}| \star \phi_J} \\
\downarrow & & \\
||f \star \psi_{j_1,\ga_1}| \star \psi_{j_2,\ga_2}| & \rightarrow & 
\boxed{ ||f \star \psi_{j_1,\ga_1}| \star \psi_{j_2,\ga_2}| \star \phi_J} \\
\downarrow & & \\
... & & 
\end{matrix}
\end{equation*}
After convolution with $\phi_J$ the output can be subsampled at intervals
$2^J$. If $f$
is an image of $N$ pixels, this uniform sampling yields a scattering 
representation $S_J f$ including a total of 
$N_J = 2^{-2J} N \sum_{n=0}^m |\Gamma|^n\binom{J}{n}$ coefficients.
The output of any wavelet convolution and modulus $|... \star \psi_{j,k}|$
can be subsampled at intervals
$2^{j-1}$ which reduces intermediate computations and barely introduces any 
aliasing. With an FFT, 
the overall computational complexity is then $O(N \log N)$.

\subsection{Scattering Metric and Deformation Stability}
\label{scatmetric}

For appropriate complex wavelets, one can prove \cite{stephane1} that
the energy $\sum_{|p|=m} \|S_J (p)f\|^2$ of a scattering layer $m$ tends
to zero as $m$ increases. This decay is fast. Numerically the maximum
network depth is typically limited to $m_0 =3$. 
 

The scattering metric is obtained with a summation over
all paths $p$:
$$\|S_{J} f - S_{J} g\|^2 = \sum_{p} \|S_J(p)f - S_J(p)g\|^2 ~,$$
where $\|S_J(p) f\|^2 = \int |S_J (p) f(x)|^2\,dx$.
Since $S_J$ is calculated by
iterating on the contractive propagator $\pavel$ (\ref{cascadeop}), it results
that it is also contractive \cite{slotine}
\[
\|S_{J} f - S_{J} g\|^2 \leq \|f - g\|^2\,.
\]
Scattering operators are not only contractive but also preserve the norm.
For appropriate complex wavelets which 
satisfy (\ref{littlewood2}) for $\delta = 0$,
one can prove \cite{stephane1} that $\|S_J f \| = \|f \|$. 

When a signal is translated $f_\tau (x) = f(x-\tau)$,
the scattering transform is also translated
\[
S_J (p) f_\tau (x) =S_J (p) f (x-\tau)
\] 
because it is computed with 
convolutions and modulus. However, when $J$ increases,
$S_J (p) f(x)$ tends to a constant because of the convolutions with $\phi_J$.
It thus becomes translation invariant and one can verify 
\cite{stephane1} that  
the asymptotic scattering metric is translation invariant:
\[
\lim_{J \rightarrow \infty} \|S_J f - S_J f_\tau \| = 0~.
\]

For classification the key scattering property is its Lipschitz
continuity to deformations $D_\tau f(x) = f(x-\tau(x))$. 
Let $|\tau|_\infty = \sup_x |\tau (x)|$ and
$|\nabla \tau|_\infty = \sup_x |\nabla \tau (x)| < 1$, where
$|\nabla \tau (x)|$ is the matrix sup norm of $\nabla \tau (x)$.
Along paths of length $|p| \leq m_0$, one can prove 
\cite{stephane1} that
for all $2^J \geq |\tau|_\infty/|\nabla \tau|_\infty$ the scattering metric
satisfies
\begin{equation}\label{lipschitz}
\|S_{J} D_\tau f- S_{J}f\| \leq C \,m_0 \,\|f\|\,
|\nabla \tau|_\infty\, \log \frac{|\tau|_{\infty}}{|\nabla \tau|_{\infty}}\,.
\end{equation}
The scattering operator is thus Lipschitz continuous to deformations,
up to a log term. It shows that
for sufficiently large scales $2^J$, the signal translations
and deformations are locally linearized by the scattering operator.

\section{Classification}\label{section_classif}

Local translation invariance and Lipschitz regularity to local
deformations linearize small deformations.
Signal classes can thus be approximated with low-dimensional
affine spaces in the scattering domain.
Although the scattering representation is implemented
with a potentially deep convolution network, learning is not
deep and it is reduced to PCA computations.
The classification is implemented with a penalized model
selection. 

\subsection{Affine Scattering Space Models}

A signal class $\cal C$ can be modeled as a realization
of a random process $F$. There are multiple sources of 
variability, due to the reflectivity of the material
as in textures, due to deformations or to various illuminations.
Illumination variability is often low-frequency and can be
approximated in linear spaces of dimension close to $10$
\cite{basrijacobs}. This property remains valid in the
scattering domain. A scattering operator also linearizes local deformations 
and reduces the variance of large classes of stationary processes.
One can thus build a linear affine space approximation of 
$S_J F$. A scattering transform $S_J F$ along progressive paths
of length $|p| \leq m_0$ is a vector of size $O(N)$, which may be 
much smaller than $N$ if $J$ is large. 

The affine space $\mathbf{A}_k$ of dimension $k$ which minimizes
the expected projection error 
$E\{\|S_J F - P_{\mathbf{A}_k}(S_J F) \|^2\}$ is
\begin{equation}\label{affinspaces}
\mathbf{A}_k = \mu_J + \mathbf{V}_{k} 
\end{equation}
where $\mu_J (p,x) =E\{S_J (p) F (x)\}$ and 
$\mathbf{V}_k$ is the space generated
by the first $k$ eigenvectors 
of the covariance operator of $S_J (p) F (x)$. 
The space dimension $k$ is limited to a maximum value $K$.

These affine space models are 
estimated by computing the empirical average and the 
empirical covariance of 
$S_J (p) f(x)$, for all training signals
$f \in \cal C$.
The empirical covariance is diagonalized
to estimate the $K$ eigenvectors of largest eigenvalues.
Under mild conditions \cite{vershynin}, the sample 
covariance matrix $ \hat{\mathbf{\Sigma}}$ 
converges in norm to the true covariance 
when the number of training signals is of the 
order of the dimensionality of the space where
$S_J F$ belongs. Dimensionality reduction is thus important to
learn affine space models from few training signals.

%

The computational complexity to estimate 
affine space models $\hat{\mathbf{A}}_{k} $ 
is dominated by eigenvectors calculations.
To compute the first $K$ eigenvectors, a
thin SVD algorithm requires
$O\left(T\, K\,  N\right)$ operations, where $T$ is the
number of training signals.

\subsection{Linear Model Selection}

Let us consider 
a classification problem with several
classes $\{\mathcal{C}_i\}_{1 \leq i \leq I}$. 
We introduce a classification algorithm which selects affine space
models by minimizing a penalized approximation error.

Each class $\cC_i$ is represented by a family of embedded
affine spaces $\mathbf{A}_{k,i}  = \hat \mu_{i} + {\mathbf{V}_{k,i}} $,
where ${\bf V}_{k,i} $ is the space generated by 
the first $k$ eigenvectors $\{e_{i,l} \}_{l \leq k}$ 
of the empirical covariance matrix $\widehat{\Sigma}_i$.
For a fixed dimension $k$, 
a space $\mathbf{A}_{k,i}$ is discriminative for $f \in \cC_i$
if the projection error of $S_J f$ in $\mathbf{A}_{k,i}$ 
is smaller than its projection in the other
spaces $\mathbf{A}_{k,i'}$:
\[
\forall i'~~,~~\|S_J f - P_{\mathbf{A}_{k,i'}}(S_{J} f)\|^2 \geq
\|S_J f - P_{\mathbf{A}_{k,i}}(S_{J} f)\|^2  ~,
\]
with
\[
\|S_J f - P_{\mathbf{A}_{k,i}}(S_{J} f)\|^2  = 
\|S_J f - \hat \mu_i \|^2  -  
\sum_{l=1}^k
|\langle S_{J} f - \hat \mu_i \,,\,e_{i,l} \rangle|^2~.
\]

Model selection for classification
is not about finding an accurate approximation model
as in model selection for regression but looks for
a discriminative model \cite{celeux}.
If $S_J f$ for $f \in \cC_i$ is close to the class centroid $\hat \mu_i$
then low-dimensional affine spaces $\mathbf{A}_{k,i}$ are highly discriminative
even if the remaining error is not negligible, because it is unlikely that any
other low-dimensional affine space $\mathbf{A}_{k,i'}$ yields a comparable error.
If $f$ is an ``outlier'' which is far from the centroid $\hat \mu_i$ then a
higher dimensional approximation space $\mathbf{A}_{k,i}$ is needed for discrimination.
One can then adjust the dimensionality of the discrimination
space to each signal $f$ by penalizing the dimension of the approximation space.
The class index $i$ of $f$ is estimated by adjusting the dimension $k$
of the space $\mathbf{A}_{k,i}$ that yields the best approximation, with a penalization 
proportional to the space dimension $k$ \cite{celeux}:
\[
\hat{\i}(f) = \argmin_{i \leq I}\min_{k\leq K}  
\|S_J f - P_{\mathbf{A}_{k,i}}(S_{J} f)\|^2 + \beta k ~.
\]

%
%

This classification algorithm depends upon
the penalization factor $\beta$ and the scale $2^J$
of the scattering transform. 
These two parameters 
are optimized with a cross-validation mechanism. 
It minimizes a classification error computed
on a validation subset of the training samples,
which does not take part in the affine model learning.

\begin{itemize}

\item Increasing the scale $2^J$ reduces the intra-class variability
of the representation by building invariance,
but it can also reduce the distance across classes.
The optimal size $2^J$ is thus a trade-off between both.
 
\item The penalization parameter $\beta$ is similar to
a threshold on
$|\langle S_J F - \hat \mu_i , e_{i,k}\rangle|^2$.
The model increases the dimension $k$ of the approximation space
if the inner product is above $\beta$. Increasing $\beta$ thus
reduces the dimension of the affine model spaces, which is needed when
the training sequence is small.
\end{itemize}

\section{Classification Results and Analysis}\label{sec_num_results}

This section presents classification
results for handwritten digit recognition,
and for texture discrimination with illumination variations. 
The scattering transform is implemented with the same
Gabor wavelets along $|\Gamma|=6$ orientations for both
problems, and
the maximum scattering length is limited to $m_0 =2$.

\subsection{Handwritten Digit Recognition}

The MNIST hand-written digit database provides a good
example of classification with important deformations. 
Table \ref{MNIST} compares scattering classification
results for training sets of variable size,
with results obtained with deep-learning convolutional networks 
\cite{ranzato_cvpr}, which currently have the best results.
Table \ref{MNIST} compares 
the PCA model selection algorithm applied on scattering coefficients
and an SVM classifier with 
polynomial kernel whose degree was optimized, also applied on scattering
coefficients.
Cross validation finds an optimal scattering scale $J = 3$, which corresponds
to translations and deformations of amplitude about $2^J = 8$ pixels, 
which is compatible with observed deformations on digits. 

Below $5\,10^3$ training examples, 
a PCA scattering classifier provides state of the art results. It yields
smaller errors 
than deep-learning convolution network which require large training sets
to optimize all network parameters with backpropagtion algorithms.
For $60\,10^3$ training samples, the deep-learning convolution network error
\cite{mnist_ranzato} is below the scattering classifier error.
Table \ref{MNIST} shows that applying a linear SVM classifier
over the scattering transform degrades
the results relatively to a PCA classifier
up to large training sets, and it requires much more computations.
This is an indirect validation of the linearization properties of the
scattering transform.

\begin{table}[t]
\caption{Percentage of error as a function of the training size for MNIST. Minimum errors are in bold. The last column gives the average model space dimension $\overline k$.}
\label{MNIST}
\begin{center}
\begin{tabular}{|c | c c c |  }
\hline
Training & ConvNets\cite{ranzato_cvpr} & Scatt+SVM & Scatt+PCA \\ 
\hline
300 & 7.18 & 21.5& \textbf{5.93} \\
1000 & 3.21& 3.06 & \textbf{2.38}\\
2000 & 2.53 & 1.87& \textbf{1.76}\\
5000 & 1.52 & 1.54 &\textbf{1.27} \\
10000 & \textbf{0.85}& 1.15& 1.2 \\
20000 & \textbf{0.76}& 0.92 &0.9\\
40000 & \textbf{0.65}& 0.85 &0.86\\
60000 & \textbf{0.53} & 0.7 &0.74 \\
\hline
\end{tabular}
\end{center}
\end{table}

\begin{figure}[h!]
\centering
\includegraphics[scale=0.35]{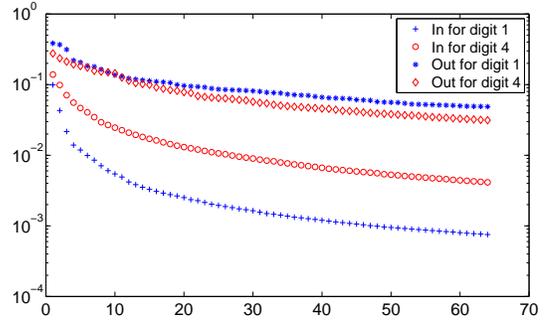}
\caption{Relative Intra-class ${\rm In}$ 
and average Outer-class ${\rm Out}$ approximation error for the digits $i=1$ and $i=4$.}\label{decay_empiric_variances}
\end{figure}

Figure \ref{decay_empiric_variances} shows the relative approximation error
when approximating a signal class with an affine model in the scattering 
domain.  
For digits $i=1$ and $i = 4$,
it gives the average Intra-class approximation error 
of $S_J F_i$ with a space $\mathbf{A}_{k,i}$ of the same class,
as a function of $k$:
\[
{\rm In(i)} = \frac
{E\{\|S_J F_i - P_{\mathbf{A}_{k,i}}S_J F_i\|^2\}}
{E\{\|S_J F_i \|^2\}}~.
\] 
It is compared with
\[
{\rm Out(i)} = \frac
{E\{\|S_J F_{i'} - P_{\mathbf{A}_{k,i}}S_J F_{i'}\|^2\, | \, i \neq i'\}}
{E\{\|S_J F_{i'} \|^2\, | \, i \neq i'\}}~.
\]
 which is the average Outer-class approximation error produced 
 by the spaces $\mathbf{A}_{k,i}$
over all samples  $S_J F_{i'}$ belonging to different classes $i' \neq i$.
The intra-class error decay is much faster than the outer-class error
decay for $k \leq 10$, which shows the discrimination ability of low dimensional
affine spaces.
For $k \geq 10$, intra-class versus outer-class distance ratio 
${\rm In}/{\rm Out}$ is approximatively $10^{-2}$ and $10^{-1}$ 
respectively for the digits $i = 1$ and $i = 4$. It
shows the discrimination power of these
affine models, and the much larger intra-class variability for 
hand-written digits $4$ than for hand-written digits $1$.

The US-Postal Service set
is another handwritten digit 
dataset, with 7291 training samples
and 2007 test images $16 \times 16$ pixels.
The state of the art is obtained with 
tangent distance kernels \cite{tangent}.
Table \ref{usps} gives 
results with a PCA model selection on scattering coefficients
and a polynomial kernel SVM classifier applied to 
scattering coefficients. The 
scattering scale was also set to $J=3$ by cross-validation.

\begin{table}[t]
\caption{Error rate for the whole USPS database.}
\label{usps}
\begin{center}
\begin{tabular}{|c c c c|}
\hline
Scatt+PCA & Scatt+SVM & Tangent kern.\cite{tangent} & humans  \\ 
\hline
2.64 & 2.64 & \textbf{2.4} & 2.37\\
\hline
\end{tabular}
\end{center}
\end{table}

\subsection{Texture classification: CUREt}

\begin{figure}[h!]
\centering
\includegraphics[scale=0.3]{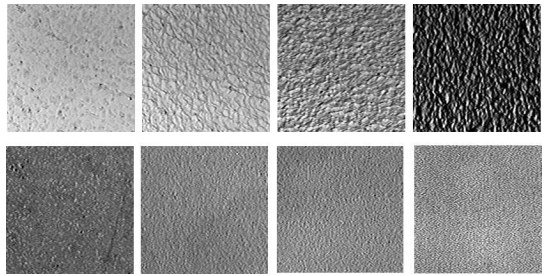}
\caption{Top row: images of the same texture material with different poses and illuminations. Bottom row: examples of textures 
that are in different classes despite their similarities.}\label{curetfig}
\end{figure}

The CureT texture database \cite{Malik} includes
61 classes of image textures of $N = 200^2$ pixels,
with 46 training samples and 46 testing samples in 
each class. Each texture class gives images of
the same material with different
pose and illumination conditions. 
Specularities, shadowing and surface normal 
variations make it challenging for classification. 
Figure \ref{curetfig} 
illustrates the large intra class variability, and also shows 
that the variability across classes is not always important.  

Classification algorithms with optimized textons
have an error rate of 5.35\% \cite{Malik} over this database,
and the best result
of 2.57\% error rate was obtained in \cite{Zisserman} with
an optimized Markov Random Field model.

Wavelets have been shown to be provide useful models for texture
analysis \cite{portilla}.
Scattering classification results are shown 
in table \ref{curet}, with exactly the same algorithm
as for digit classification. With a PCA it
greatly improves existing results with 
an error rate of $0.2\%$.
The SVM classifier with an optimized polynomial kernel on
scattering coefficients achieves a larger error rate of $1.71\%$.

\begin{table}[t]
\caption{Error rate for the CUREt database}
\label{curet}
\begin{center}
\begin{tabular}{|c c c c|}
\hline
Scatt+PCA & Scatt+SVM & Textons \cite{Malik} & MRFs \cite{Zisserman} \\ 
\hline
$\bf{0.2 \pm 0.08}$ & 1.71 & 5.35 & 2.57 \\
\hline
\end{tabular}
\end{center}
\end{table}

The cross-validation adjusts the scattering scale $2^J = 2^7$
which is the maximum value. Indeed, these textures are
fully stationary and increasing the scale reduces the variance of 
the scattering coefficients variability across realizations.
Scattering vectors $S_J f$ at large scales $2^J$
have a small stochastic 
variability within each texture class because of the averaging
by $\phi_J$. Moreover, global invariance
to rotation and illumination changes is provided by the PCA classification
algorithm. These invariant linear space models
are learnt effectively even with few training samples.
This example shows that
linear models are a simple yet powerful 
mechanism to generate invariance for classification problems.

\section{Conclusion}

As a result of their translation invariance and Lipschitz regularity to
deformations,
scattering operators provide appropriate representations to model complex
signal classes with affine spaces calculated with a PCA. 
Classification with model selection provides state
of the art results with limited training size sequences, for handwritten
digit recognition and textures. As opposed to discriminative classifiers such
as SVM and deep-learning convolution networks, these algorithms 
learn a model for each class independently from the others, which leads to fast 
learning algorithms. 

Scattering operators can be defined on more general Lie groups other
than the group of translations, such as the group of rotations or scaling \cite{stephane1}. 
The intra-class variability
due to the action of several transformation groups can be contracted
by combining scattering operators adapted to each of these groups \cite{stephane1}.
On signal classes including clutter and more complex variability, 
one can estimate the deformation group responsible of most of 
the intra-class variability, provided one has enough training samples.

\end{document}